\documentclass[conference]{IEEEtran}
\IEEEoverridecommandlockouts
\usepackage{cite}
\usepackage{amsmath,amssymb,amsfonts}
\usepackage{algorithmic}
\usepackage{graphicx}
\usepackage{textcomp}
\usepackage{xcolor}
\usepackage{subcaption}
\usepackage{multirow}
\usepackage{url}
\usepackage{booktabs}
\def\BibTeX{{\rm B\kern-.05em{\sc i\kern-.025em b}\kern-.08em
    T\kern-.1667em\lower.7ex\hbox{E}\kern-.125emX}}

\begin{document}

\title{Infrastructure-less Wireless Connectivity for Mobile Robotic Systems in Logistics: Why Bluetooth Mesh Networking is Important?}

\author{
\IEEEauthorblockN{Adnan Aijaz}
\IEEEauthorblockA{
\text{Bristol Research and Innovation Laboratory, Toshiba Europe Ltd., Bristol, United Kingdom}\\
adnan.aijaz@toshiba-bril.com}
}    

\maketitle

\begin{abstract}
Mobile robots have disrupted the material handling industry which is witnessing radical changes. The requirement for enhanced automation across various industry segments often entails mobile robotic systems operating in logistics facilities with little/no infrastructure. In such environments, out-of-box low-cost robotic solutions are desirable. Wireless connectivity plays a crucial role in successful operation of such mobile robotic systems. A wireless mesh network of mobile robots is an attractive solution; however, a number of system-level challenges create unique and stringent service requirements. The focus of this paper is the role of Bluetooth mesh technology, which is the latest addition to the Internet-of-Things (IoT) connectivity landscape, in addressing the challenges of infrastructure-less connectivity for mobile robotic systems. It articulates the key system-level design challenges from communication, control, cooperation, coverage, security, and navigation/localization perspectives, and explores different capabilities of Bluetooth mesh technology for such challenges. 
It also provides performance insights through real-world experimental evaluation of Bluetooth mesh while investigating its differentiating features against competing solutions. 
\end{abstract}

\begin{IEEEkeywords}
Automation, Bluetooth, cooperative robotics, control, distributed, mesh, mobility, logistics, swarm, warehouse.
\vspace{-1.5em}
\end{IEEEkeywords}

\section{Introduction}
Automation is the key to improving efficiency in material handling operations. Mobile robotic systems for automation in logistics and warehousing is a growing industry segment which is expected to become a \$81bn market by 2030 \cite{report_log_wh}. Enhanced automation via mobile robots is also seen as an enabler of \emph{beyond 4.0} industrial systems \cite{TI_PIEEE}. The need for mobile robots is also accelerating across various industry segments for combating COVID-19 disruption \cite{mmh_covid}. 

Mobile robotic systems employed by e-commerce giants like Amazon, Alibaba and Ocado are purpose-built for the environment they work in. This requires necessary infrastructure for the system to operate smoothly and efficiently. However, the material handling industry is rapidly evolving with increased penetration of mobile robotic systems in logistics facilities with little or no infrastructure. Prominent examples include loading/unloading of delivery trucks, goods-to-person delivery systems (in retail, construction, healthcare, etc.), makeshift warehouses, baggage storage and retrieval systems, and luggage/cargo transport at airports. In such ad-hoc environments, out-of-box and low-cost robotics solutions are desirable for flexible operation. A team of small low-cost robots operating in a cooperative manner provides a viable solution. However, such ad-hoc environments may not always have a global communication/connectivity infrastructure (e.g., access points or base stations).

Wireless connectivity for mobile robotic systems operating in such ad-hoc environments is characterized by unique service requirements. It must be transparent to frequent topology changes arising from mobility. It must be scalable without any centralized coordination or  global network state. It should also have the capability of handling different communication patterns (unicast, multicast, group, etc.) arising from task-oriented or management-oriented information exchange. Besides, it must facilitate self-deployment, cooperative task management, and autonomous navigation capabilities. A wireless mesh network of mobile robots is an attractive solution. However, such a network must fulfil these service requirements while providing flexible, versatile, scalable, and reliable connectivity.

Bluetooth mesh is the latest addition to the Internet-of-Things (IoT) connectivity landscape. Released in 2017, the Bluetooth mesh specification \cite{BT_mesh} provides a full-stack solution for mesh networking. It provides a simple, flexible, robust, secure, and efficient wireless mesh networking solution. It implements a completely decentralized architecture that provides connectivity for potentially thousands of devices in a completely connection-less manner. Concurrent with the development of Bluetooth mesh was the Bluetooth 5.0 standard \cite{BT_5} which provides various enhancements including range extension, higher data rates, and new data transfer modes (advertising extensions). Bluetooth mesh has been adopted for various IoT applications including home/building automation, smart lighting, and industrial wireless sensor networks. 

\subsection{Related Work}
Ad-hoc and mesh networking paradigms  have been widely recognized as promising solutions for communication in mobile robotic networks \cite{MANET_ref,mesh_no,rob_mesh_sensor}. Various studies (e.g., \cite{route_rob1,route_rob2,route_rob3}) have investigated routing protocols for mobile robot teams. However, the role of flooding-based communication techniques for mobile robotic networks remains unexplored. 

Recent studies have identified the need of wireless connectivity for improving the performance of collaborative robotics systems when operating in distributed manner \cite{wireless_collab,collab_multi}. Swarm robotics techniques, exploiting wireless connectivity,  for collaborative multi-robot systems have also appeared in literature \cite{swarm_rob1, swarm_rob2, swarm_rob3}. Collaborative mobile robotics testbeds are also emerging \cite{umb_rob}. However, system-level design challenges of cooperative and out-of-box robotic systems operating in ad-hoc logistics environments have not been covered in prior art. 

The performance of Bluetooth mesh has been the focus of some recent studies \cite{und_perf_BT_mesh,BT_mesh_analysis,exp_BT_mesh_2}.  However, its role for addressing the challenges of mobile robotic systems in logistics has not been investigated before. Besides, a number of gaps exist with respect to its real-world experimental evaluation.

\subsection{Key Contributions}
The main objective of this paper is to investigate the role of Bluetooth mesh in addressing the challenges of infrastructure-less wireless connectivity for mobile robotic systems in logistics. Our key contributions are as follows. 
\begin{itemize}
\item We investigate the key connectivity requirements of mobile robotic systems operating in ad-hoc logistics environments and discuss system-level design challenges from communication, control, cooperation, coverage, security, and navigation/localization perspectives. 

\item We discuss different technology features of Bluetooth mesh for addressing these system-level challenges.

\item We conduct experimental evaluation of Bluetooth mesh, based on a testbed of Nordic nRF52840 devices (https://www.nordicsemi.com/Software-and-Tools/Development-Kits/nRF52840-DK), from the perspective of different system-level challenges. 

\item We provide a comparison of Bluetooth mesh and other potential mesh technologies (for mobile robotic systems) including routing-based approaches and recent synchronous flooding techniques.
\end{itemize}

\begin{figure}
\centering
\includegraphics[scale=0.22]{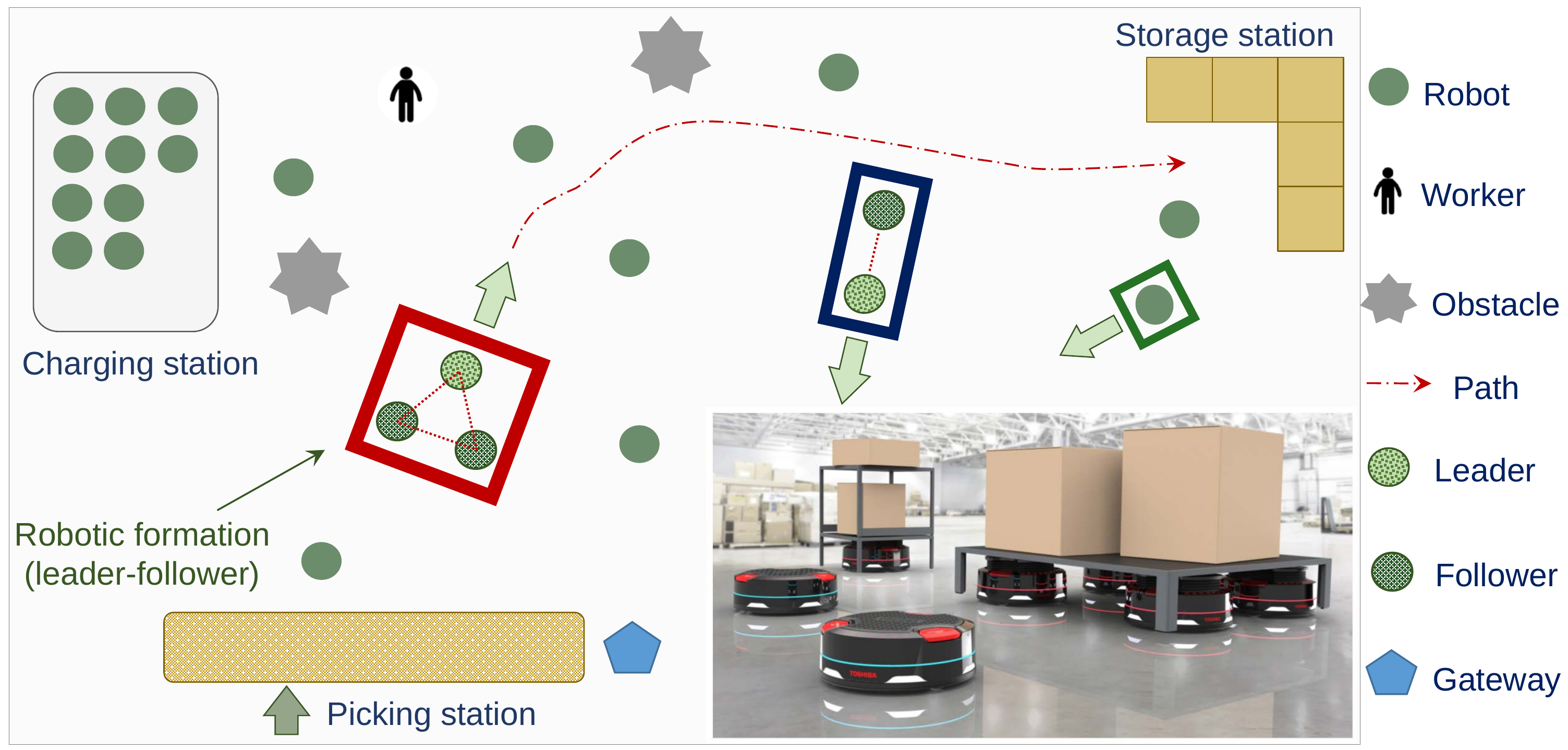}
\caption{Illustration of a mobile robotic system in material handling.}
\label{rob_sys}
\vspace{-1.5em}
\end{figure}

\section{Characterizing Connectivity Requirements}
In order to characterize the connectivity requirements of mobile robotic systems for ad-hoc logistics environments, we discuss the operational aspects of such systems. The logistics and warehousing sector uses different types of robotic systems; however,
our focus is autonomous mobile robots that transport inventory and materials throughout a logistics facility. Unlike automated guided vehicles (AGVs), that rely on fixed paths guided by wires or magnetic tapes, autonomous mobile robots navigate more flexible routes based on the environment.

Fig. \ref{rob_sys} illustrates a mobile robotic system for material handling operations. For the sake of characterizing connectivity requirements, we consider the scenario of a makeshift warehouse where variable-sized boxes/pallets are transported via a team of small and low-cost mobile robots which are deployed in an out-of-box manner.   In such a scenario, there is minimal infrastructure. Typically, robots transport items from a picking area to a storage area. There can be a dedicated charging station where robots charge their batteries. Cooperation is of utmost importance for such mobile robots as it improves  efficiency and performance of transport operations. Robots may individually transport smaller items; however, for larger items, \emph{collective transport} is necessary where multiple robots collaboratively transport an item while acting in a certain formation. Such a robotic formation often involves a leader robot and multiple follower robots. Overall, for such a mobile robotic system, following connectivity requirements (which are similar across various use-cases) must be fulfilled. 
\begin{itemize}
\item Transparency to topological changes due to mobility. 
\item Support for heterogeneous communication patterns arising from task-centric, cooperation-centric, control-centric, and management-centric  information exchange. 
\item Reliable delivery of information for successful operation of automation tasks. 
\item Flexibility for collective transport and other tasks. 
\item Support for distributed or hybrid control architectures.
\item Scalable operation for any number of robots with minimal/no overheads. 
\item Security to prevent hijacking and malicious attacks. 
\item Support for navigation and localization capabilities. 
\end{itemize}


%
%
%
%
%
%
%
%

\section{Bluetooth Mesh -- A Brief Overview}
\subsection{Protocol Stack and Architecture}
\begin{figure}
\centering
\includegraphics[scale=0.28]{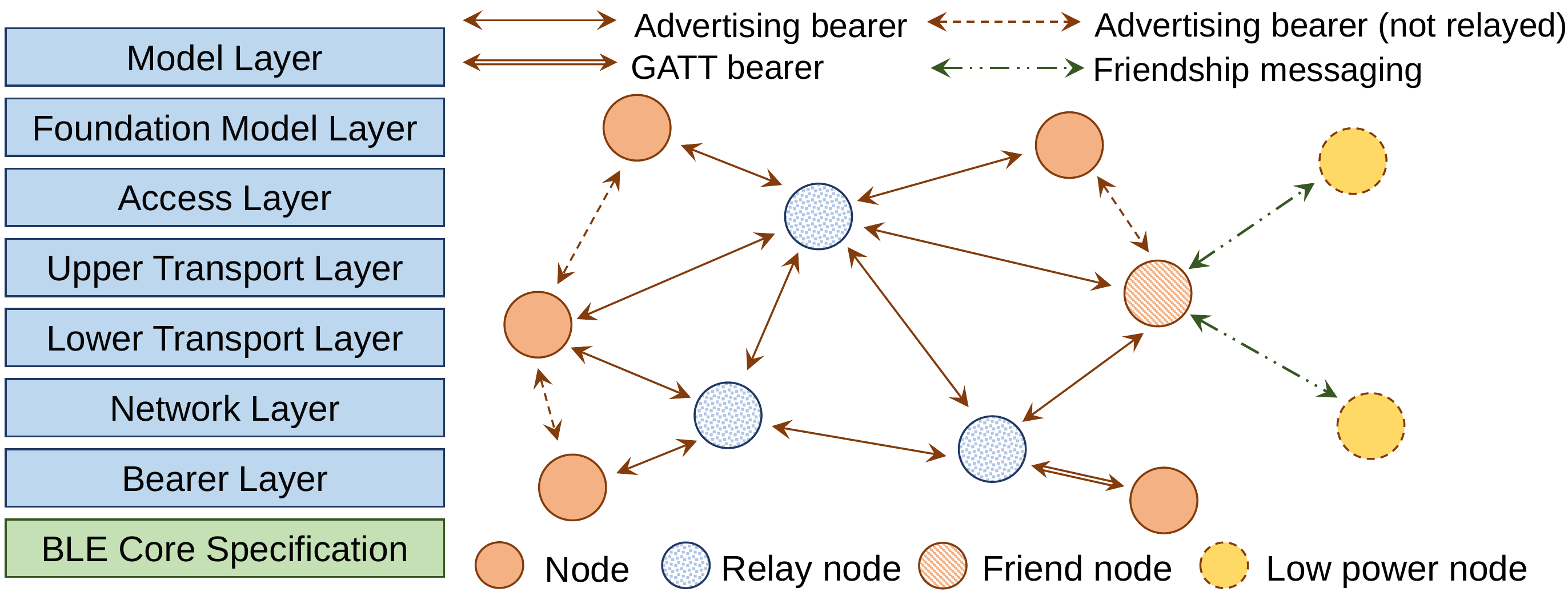}
\caption{Protocol stack and architecture of a Bluetooth mesh network.}
\label{arch_stack}
\vspace{-1.5em}
\end{figure}

The protocol stack of Bluetooth mesh is shown in Fig. \ref{arch_stack}. Bluetooth mesh is built on top of the Bluetooth low energy (BLE) standard, sharing the same Physical layer and the link layer.  The function of other layers is described as follows.

The \emph{bearer layer} defines how mesh protocol data units (PDUs) are handled by the link layer. Two bearers have been defined: an \emph{advertising} bearer that exploits BLE advertising and scanning features to transmit and receive mesh PDUs, and a \emph{generic attribute profile (GATT)} bearer that allows a device lacking Bluetooth mesh support to indirectly communicate with the mesh network. 
The \emph{network layer} defines address types and message formats that allow upper layer PDUs to be transported by the bearer layer. It also implements relay and proxy features. The \emph{lower transport layer} provides segmentation and reassembly functions for mesh PDUs. It also provides acknowledged/unacknowledged delivery of messages. 
The \emph{upper transport layer} handles authentication, encryption and decryption data coming to/from the access layer. It is also responsible for transport control messages (friendship, heartbeat, etc.) exchanged between peer nodes. 
The \emph{access layer} defines how higher layer applications utilize the layer below (upper transport layer). It defines the format of application data. It also defines and controls the encryption/decryption process at the upper transport layer. 
The \emph{foundation model layer} is responsible for implementation of models  required for configuration and management of a mesh network. 
The \emph{model layer} deals with implementation of models and basic node functionality (behaviors, states, state bindings, and messages) in standard or customized application scenarios. There are three types of models: server, client, and control. 


Fig. \ref{arch_stack} also captures the system architecture of a Bluetooth mesh network. Devices which are part of the mesh network are known as \emph{nodes}. A device joins the mesh  network by the \emph{provisioning} process \cite{BT_mesh}. All nodes can receive and transmit  messages; however,  a node may have additional capabilities. 
The \emph{relay nodes} are able to retransmit received messages over the advertising bearers. Through relaying, a message can traverse the entire multi-hop mesh network. Relaying of messages in the mesh network can be optimized through time to live (TTL) values and cache-based forwarding. 
The \emph{low power node} (LPN) has limited power resources and must save as much energy as possible. Such a node operates in a mesh network with significantly reduced duty cycle. LPNs predominantly operate as senders, although they can occasionally receive messages. 
The \emph{friend node} assists operation of LPNs in the mesh network. It stores messages destined to LPN and forwards upon request based on a polling-based mechanism. 
The \emph{provisioner node} is capable of adding a device to the mesh network via the provisioning process. 
Note that some nodes have multiple constituent parts, each of which can be individually controlled. These parts are known as \emph{elements}. 

\subsection{Managed Flooding}
\begin{figure}
\centering
\includegraphics[scale=0.29]{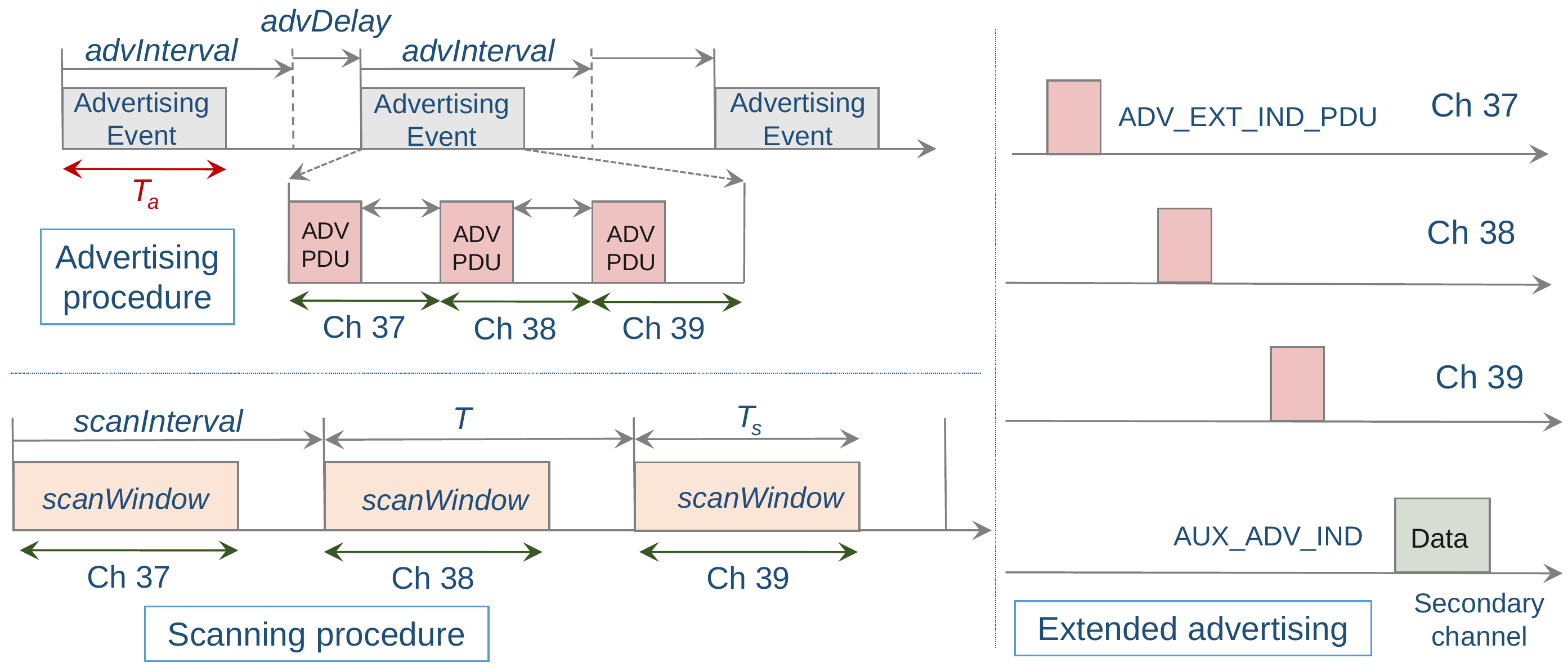}
\caption{Illustration of advertising and scanning procedures in Bluetooth mesh.}
\label{adv_scan}
\vspace{-1.5em}
\end{figure}

Bluetooth mesh adopts \emph{managed flooding} to deliver messages. The managed flooding mechanism is completely asynchronous and provides a simple approach to propagate messages in the mesh network using broadcast. A transmitted message is potentially forwarded by multiple relay nodes. Message transmissions  predominantly take place via the advertising bearer. A source node injects its message in the mesh network through an advertising process. An advertising event is a cycle of advertising operations where mesh protocol data units (PDUs) are transmitted in sequence over each of the three (primary) advertising channels (i.e., channel 37, 38 and 39). Multiple advertising events can be configured at the network layer to improve the reliability of message injection. The time between two advertising events is dictated by the advertising interval (\emph{advInterval}) and a random advertising delay (\emph{advDelay}).
The relay nodes scan the advertising channels and listen to the advertising information of the neighbors. The scanning operation is performed in scanning events that repeat after scanning interval (\emph{scanInterval}). Multiple relays scanning on different advertising channels at different times increase the probability of message propagation in the mesh network. The advertising/scanning procedures are illustrated in Fig. \ref{adv_scan}. 

The managed flooding approach provides multi-path diversity that improves reliability. However, it can also increase collisions on the advertising channels. Bluetooth mesh offers various configuration options to overcome this issue, for example TTL limitations, message cache to restrict forwarding, and  random back-off periods between different advertising events and different transmissions within an advertising event. Bluetooth mesh implements a publish/subscribe messaging paradigm. Publishing refers to the act of sending a message. Typically, messages are sent to unicast, group or virtual addresses. Nodes can be configured to receive messages sent to specific addresses which is known as subscribing. 

Bluetooth 5.0 introduces an \emph{extended advertising} mode (illustrated in Fig. \ref{adv_scan}) that exploits additional channels for data transmission. A source node transmits short advertising indication PDUs (on primary advertising channels) which include a pointer to a secondary advertising channel (randomly selected from the remaining 37 BLE channels) over which data transmission takes place. Extended advertisements provide the capability of transmitting more data than that allowed on legacy advertisements. Moreover, data transmissions can exploit uncoded (1 Mbps and 2 Mbps) and coded (500 kbps and 125 kbps) Physical (PHY) layers. Extended advertisements also enable periodic advertisements which allow broadcasting of data to (unconnected) mesh nodes at fixed intervals.

%

\section{Tackling System-level Design Challenges}
Successful operation of a mobile robotic systems for material handling operations in an out-of-box manner is largely dependent on addressing various system-level design challenges. Wireless connectivity plays a crucial role in tackling these challenges. We discuss various design challenges and the role of Bluetooth mesh in addressing these challenges. 

\subsection{The Communication Challenge} 
One of the most important challenges for infrastructure-less wireless connectivity for mobile robotic systems is handling of versatile communication requirements. 

\textit{Mobility Support} -- The connectivity technology must be transparent to mobility of mobile robots. The communication requirements need to be fulfilled without any disruption due to mobility. Bluetooth mesh \emph{natively} supports mobility due to its managed flooding approach which is transparent to topology changes arising from node mobility and does not require maintaining network state in terms of routing/forwarding tables. 

\textit{Heterogeneous Traffic} -- The connectivity technology must be capable of handling heterogeneous traffic, in terms of communication patterns, varying payloads, message frequencies, etc. Such heterogeneity arises due to task-oriented, management-oriented, navigation-oriented, and cooperation-oriented communication in a network of mobile robots. Bluetooth mesh has native support for unicast and multicast communication and it can simultaneously support different traffic patterns such as one-to-all, one-to-one, one-to-many, many-to-one, many-to-one, and all-to-all. Different types of Bluetooth mesh bearers can be leveraged to simultaneously handle different types of traffic like event-triggered, periodic, aperiodic and time-triggered.

\textit{Dynamic Environments} -- Mobility is not the only factor characterizing dynamic environments. A mesh network of mobile robots can dynamically change with addition/removal of some robots or when robots are engaged in certain activities like battery charging or operating in a formation. The connectivity technology must ensure minimal reconfiguration of other robots in the network in such dynamic environments. Due to its completely decentralized approach, Bluetooth mesh does not require reconfiguration of other network nodes in case of removing, replacing or adding new nodes in the network. The publish/subscribe model \cite{BT_mesh} and the use of group and virtual addresses \cite{BT_mesh_overview} provides further support for reconfiguration-free operation in dynamic environments.

\textit{Flexible Connectivity} -- Flexibility of wireless connectivity is crucial for successfully handling different types of tasks (e.g., collective/individual transport, storage/retrieval, navigation assistance, and pattern formation) and requirements (e.g., mobility and traffic heterogeneity). Bluetooth mesh offers  much-needed flexibility for infrastructure-less connectivity. It does not require any  centralized management like routing and scheduling. It is free from time synchronization overheads. It natively supports segmentation and reassembly as well as acknowledged/unacknowledged messages without  complex protocol-level design or enhancements. It has the area isolation capability, i.e., parallel operation of multiple subnets. This is particularly important as multiple robotic teams/formations can simultaneously operate in different parts of the logistics facility. It also has native support for publish/subscribe architecture which is important for message dissemination based on widely-used robot operating system (ROS). Last but not least, different node and address types ensure that robots can simultaneously engage in different types of tasks.

\textit{Scalability} -- Out-of-box mobile robotic systems can be deployed for small to large scale logistics applications. The completely decentralized flooding approach of Bluetooth mesh can be scaled to a large number of nodes. Bluetooth mesh can support up to 32767 nodes in a network over 126 hops \cite{BT_mesh}. 

\textit{Interoperability} -- Interoperability of a communication technology prevents single vendor lock-in which can be an issue for industry verticals like logistics. Bluetooth mesh provides multi-vendor interoperability; hence, devices from different manufacturers can work together in a mesh network of robots.

\subsection{The Cooperation Challenge}
Cooperation is at the heart of low-cost and out-of-box mobile robotic systems for material handling. It improves overall performance and efficiency of material handling tasks. From a system-level perspective, the connectivity layer must handle cooperation-centric requirements in a seamless manner. Bluetooth mesh provides various features which are promising for realizing cooperation in mobile robotic systems. Material handling operations typically involve transportation of large objects across logistics facilities which can be most efficiently handled through \emph{collective transport} that entails multiple robots acting collaboratively in a certain geometric formation. Formation control of mobile robots is most efficiently handled through a leader-follower approach. Bluetooth mesh facilitates communication exchange for \emph{formation synthesis} with complete flexibility as multiple formations can be simultaneously created and operated in different parts of the logistics facility. By leveraging advertising and scanning procedures, a best-fit robot among available robots can be elected as leaders for a certain task. Similarly, a leader robot can recruit follower robots for a collective transport task.  

Once a robotic formation is synthesized, it must be maintained during collective transport. Communications-based control of a robotic formation, in the form of single-hop connectivity between a leader and multiple follower robots, provides a simple and efficient solution \cite{comms_formation}. Such communication demands enhanced timeliness capabilities for accurate path-tracking between the leader and the followers. As Bluetooth mesh stack is built on BLE standard, connection-oriented communication \cite{BT_5} can be employed within a robotic formation for improved timeliness. An alternate solution is periodic advertisements of Bluetooth mesh which allow synchronized communication without connection mode operation. 

Through virtual and group addresses \cite{BT_mesh_overview}, robots can be seamlessly added/removed from a formation. Bluetooth mesh models \cite{BT_mesh}, which define set of states, state transitions, state bindings, and associated behaviors for every element in a node, are particularly attractive for defining defining all the functionality associated with cooperation-centric tasks. For example, a customized model can be defined for formation synthesis task that exploits different states (elected leader, recruit followers, selected follower, etc.) of robots.

\subsection{The Control Challenge}
Conventional infrastructure-based logistic facilities employ a centralized control architecture for system management. However, centralized architectures require complex designs and may not provide the required flexibility for ad-hoc logistics. For out-of-box and cooperative robotics, distributed control architectures are desirable for flexible operation and adaptability to user requirements. Moreover, distributed control solutions can leverage simple swarm intelligence algorithms for self-organized behavior. 

Bluetooth mesh provides a completely decentralized  architecture which is promising for realizing distributed control algorithms. 
In the smart lighting community, Bluetooth mesh is widely recognized as the \emph{killer of the lighting switch} as it underpins distributed control capabilities. The decentralized connectivity architecture and the native support for heterogeneous traffic provides the capability of local decisions at robots as required by distributed control and/or swarm algorithms. Such local decision can be related to task delegation, storage, transport, formation control, retrieval, navigation, etc. 

The group communication capabilities and the native support for different traffic patterns also provides the capability of a hybrid control architecture where certain management aspects like task allocation can be handled in a centralized manner (e.g., through cloud-based agents) while other aspects can be handled in a distributed manner. Bluetooth mesh also allows humans to directly interact (via the GATT bearer) with a mesh network of mobile robots.

\subsection{The Security Challenge}
Security of mobile robotic systems  for logistics is crucial to prevent attacks that can lead to  service disruption. The low-cost and out-of-box nature of mobile robotic systems in question create various additional security vulnerabilities arising from resource constraints, use of decentralized algorithms, and the possibility of physical tampering. Such vulnerabilities can potentially result in loss of confidentiality and availability.

Conventional wireless protocols typically consider security as an afterthought rather than an upfront design challenge. In contrast, security is mandatory in Bluetooth mesh. It implements a multi-layer security approach which spans the network, the devices, and the applications \cite{BT_mesh_overview}. The keys for network security, application security, and device security are completely independent. All messages in a Bluetooth mesh network are encrypted (twice) and authenticated. Devices join the mesh network based on a secure provisioning process. These features ensure confidentiality of mesh communication while preventing eavesdropping attacks on robots.

Bluetooth mesh uses a secure process to remove nodes from a network which prevents trashcan attacks. If a robot becomes faulty and needs to be disposed of, it can be added to a black list after which a key refresh procedure is initiated. Hence, compromised robots cannot be added back to the network. 

Bluetooth mesh has protection against replay attacks which can be carried out through intercepting messages and retransmitting them later with the goal of tricking recipients in conducting unauthorized actions. Such attacks can be detrimental for distributed control functions and local sensing decisions by robots. The basis for protection against replay attacks is the use of two fields in mesh PDUs: the sequence number and the initialization vector. A receive message is discarded if these values are less than or equal to the previous PDU.

\begin{figure}
\centering
\includegraphics[scale=0.25]{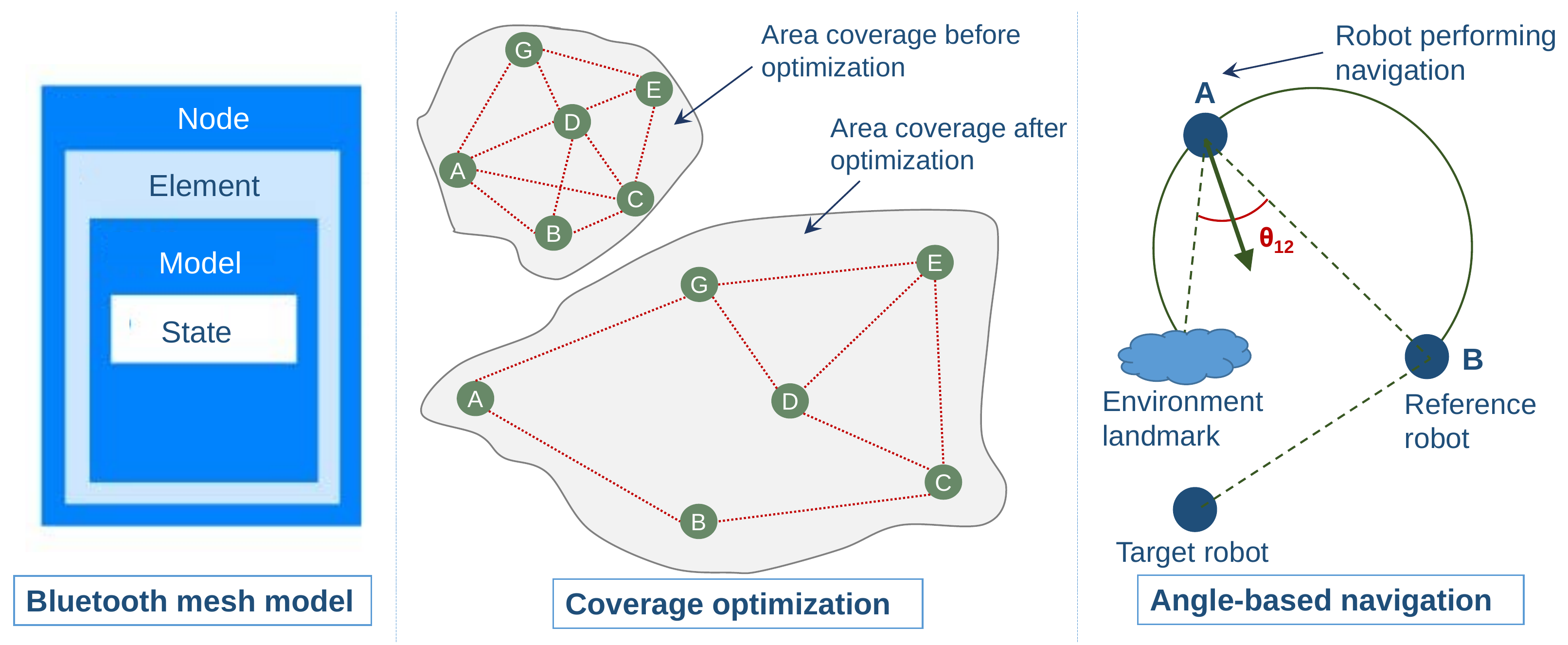}
\caption{Mesh models, coverage optimization, and angle-based navigation.}
\label{mesh_cap}
\vspace{-1.6em}
\end{figure}

\subsection{The Coverage Challenge}
A key challenge in realizing material handling operations with mobile robots under no global communication infrastructure is to maintain complete network connectivity under movement constraints. Conventional techniques for coverage optimization in purpose-built logistics environments rely on deployment of access points. In a wireless mesh network of mobile robots, coverage optimization (illustrated in Fig. \ref{mesh_cap}) must be achieved without centralized coordination or complex path-planning under global network state.

Bluetooth mesh is promising for addressing the challenges of coverage optimization in mobile robotic systems. It underpins completely independent and distributed control decisions for optimizing coverage. In particular, the advertising and scanning procedures  provide a simple and pragmatic solution for maintaining neighborhood information. During advertising phase, a node can transmits beacons in one or more advertising events. During the scanning phase, a node scans for beacons from neighbouring nodes. These beacons can be used as indicator of the received power (RSSI). For mutual connectivity, a node can piggyback the received RSSI from a node on the transmitted beacon along with the ID of the node. Based on the neighborhood information, a node can determine the state of its local connectivity and if that needs to be optimized through movement. Native support for event-triggered traffic in Bluetooth mesh ensures that coverage optimization can be performed in conjunction with other tasks. 

\subsection{The Navigation/Localization Challenge}
Navigation and localization are key aspects of any mobile robotic system. The connectivity layer must fulfil the requirements of navigation and localization while providing other functionalities. Navigation can be achieved in a completely autonomous manner or based on some guidance control techniques (e.g., line following based on magnetic tapes or wires), and it is essential for various tasks including transport and coverage optimization. Localization is important for a number of tasks including formation synthesis and tracking of pallets/objects. Relative localization is more important than absolute localization for cooperative robotics applications. 

In practice navigation is possible without localization. Bluetooth mesh provides the capability of connectivity-navigation co-design. Through integration of visual sensing (using onboard cameras) and distributed mesh communication, simple angle-based control techniques (illustrated in Fig. \ref{mesh_cap}) are possible where a robot uses environment landmarks or other robots as reference and navigates in the environment for coverage optimization, picking/storing objects, etc. Bluetooth mesh is also promising for cooperative navigation techniques \cite{coop_nav} for swarm robotics through direct/indirect communication. 

Bluetooth 5.1 comes with radically improved radio direction finding (RDF) capabilities based on angle-of-arrival (AoA) and angle-of-departure (AoD) techniques \cite{BT_RDF}. Integration of RDF and mesh communications provides a promising connectivity-localization co-design approach for relative localization as required for formation synthesis and other cooperative tasks. 


\begin{figure}
\centering
\includegraphics[scale=0.25]{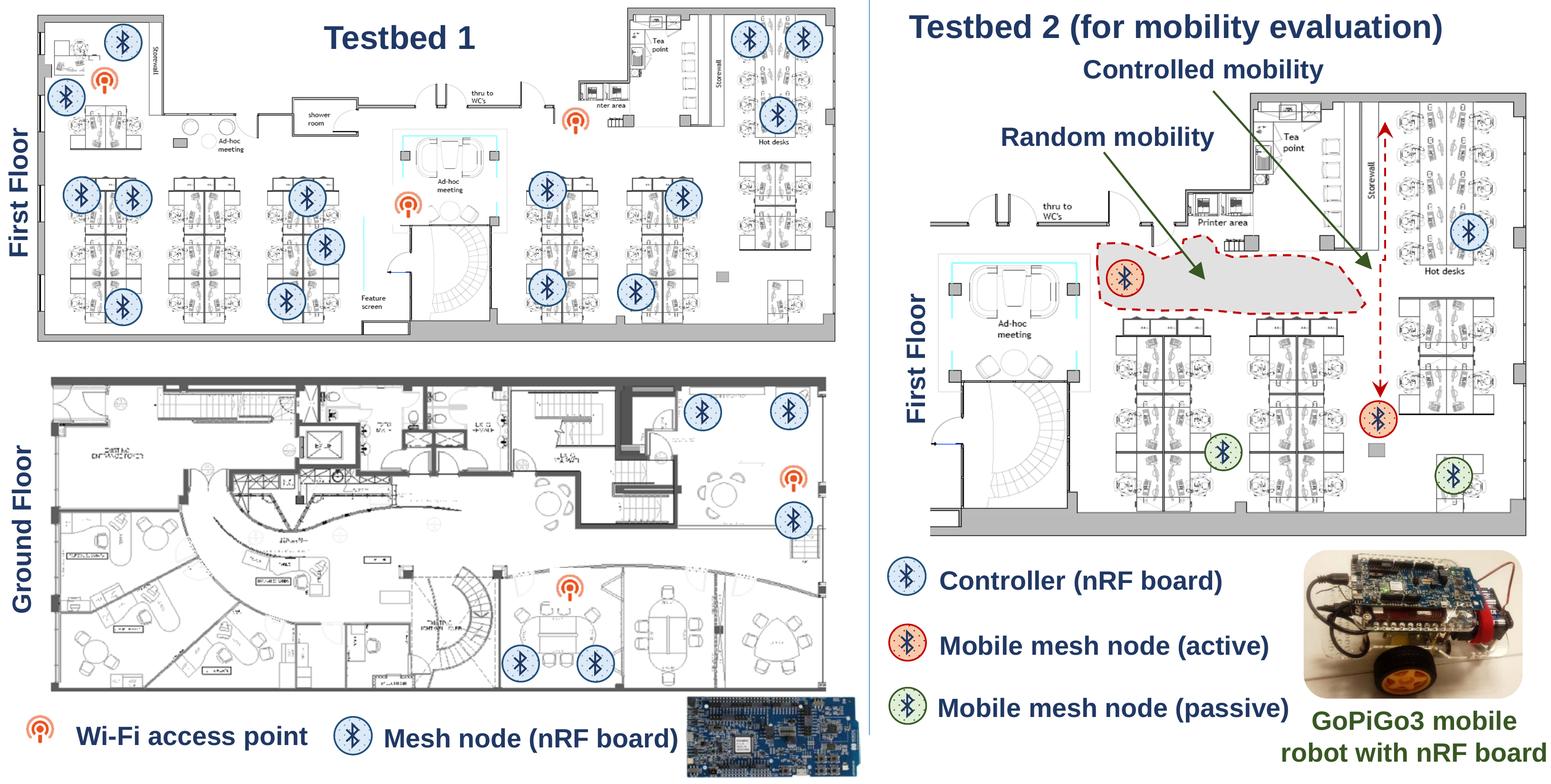}
\caption{Schematic of the two Bluetooth mesh testbeds used for evaluation.}
\label{testbed_setups}
\vspace{-1.6em}
\end{figure}

\section{Performance Evaluation}
We have conducted experimental evaluation of Bluetooth mesh with three key objectives: (i) its performance from the perspective of communication, control, and cooperation challenges, (ii) its customizability through parametric adaptation, and (iii) its viability for mesh networks under mobility. Fig. \ref{testbed_setups} shows the two testbeds used in our evaluation. In the first testbed, we have deployed 20 nRF52840  boards over two floors of our office building (approximately 600
sq. meters). The testbed setup, which is stretched over a maximum of 4 hops, depicts a challenging multi-hop mesh scenario due to weak link between the two floors. The testbed nodes experience interference (from Wi-Fi access points and other Bluetooth devices in office). We use the Nordic nRF5 software development kit (SDK) for evaluation. The testbed nodes are time synchronized via the precision time protocol (PTP) for latency measurements. The default parameters are as follows: transmit power of 0 dBm, advertising interval (\emph{advInterval}) of 20 millisecond (ms), scanning interval (scanInterval) of 2000 ms, message size of 11 bytes. All nodes are configured as relays and the number of advertising events is set to 3 and 2 for source and relay nodes, respectively. The results are repeated over 100 iterations with a new message every 1000 ms.

\subsection{Performance of Group Communication Capabilities}
\begin{figure}
\centering
\includegraphics[scale=0.27]{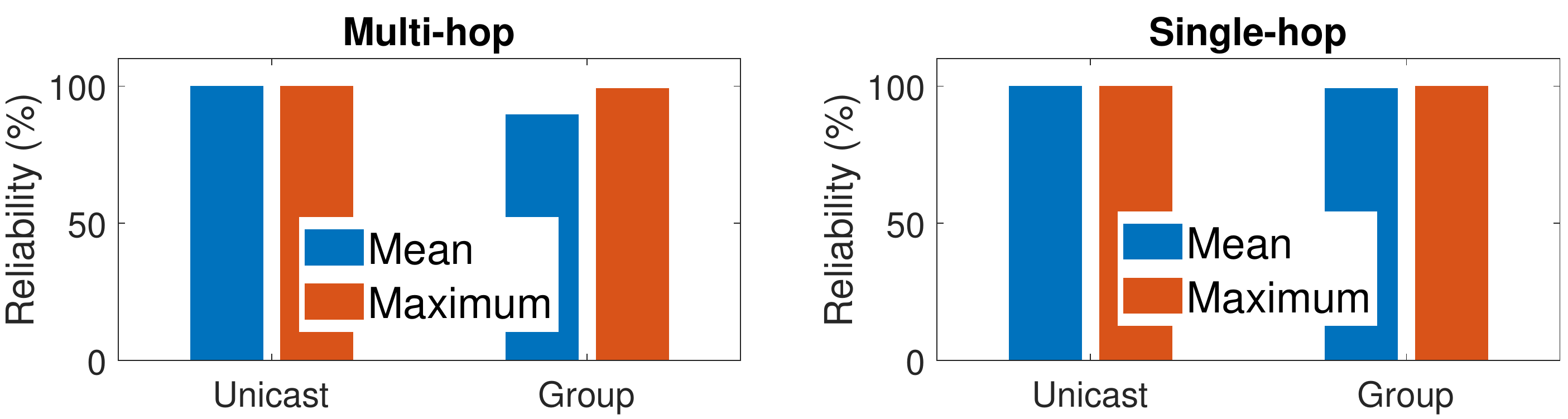}
\caption{Reliability performance of unicast and group modes.}
\label{rel_ug}
\vspace{-1.4em}
\end{figure}

\begin{figure}
\centering
\includegraphics[scale=0.27]{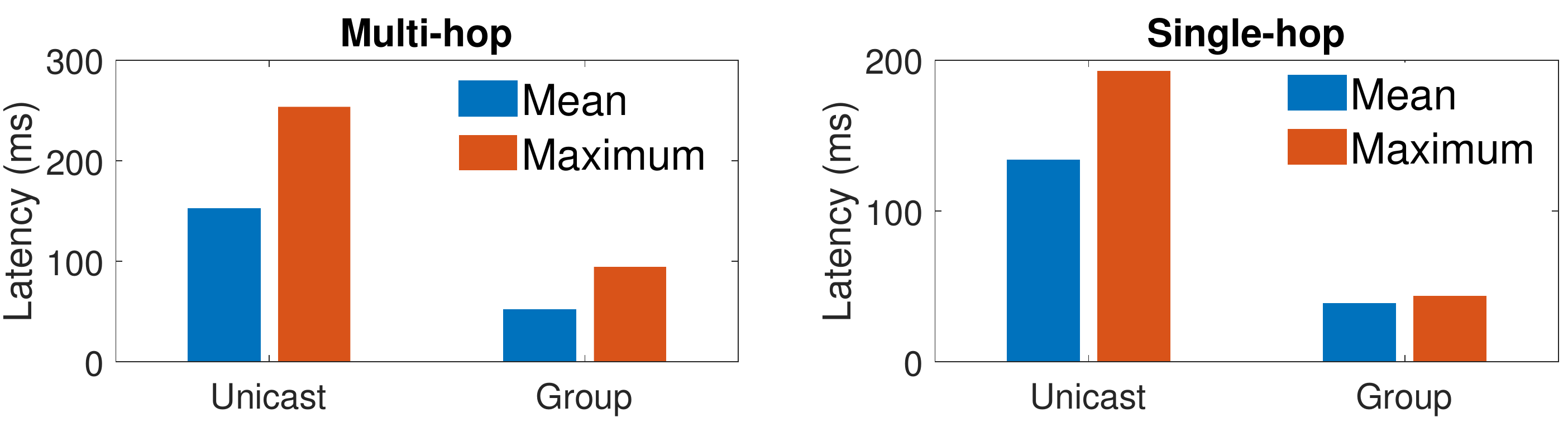}
\caption{Latency performance of unicast and group modes.}
\label{lat_ug}
\vspace{-1.4em}
\end{figure}

Group communication is important for various system-level challenges. It entails one-to-many and many-to-one communication patterns. Bluetooth mesh offers two different modes  for group communication: unicast mode and group mode. We evaluate the latency and reliability  of these modes under two different groups: a multi-hop group consisting of one controller and 14 slave nodes (spread over both floors), and a single-hop group of one controller and 7 slave nodes (left part of first floor). In both groups, the controller sends a command message to all slaves  which send an acknowledgement upon reception. In unicast mode, the controller sends a message to each node individually. The message is acknowledged; hence, the controller keeps sending until an acknowledgement is received. In the group mode, the controller sends a group message to all slaves. This message is also acknowledged; however, it is only sent twice
(i.e., over two advertising events). The reliability (packet delivery ratio) and round-trip latency  results are shown in Fig. \ref{rel_ug} and Fig. \ref{lat_ug}, respectively. The unicast mode achieves 100\% reliability in both groups. The reliability of group mode is affected by fixed  message transmissions. It achieves 89.6\% and 99.07\% reliability in multi-hop and single-hop groups, respectively.  The group mode outperforms the unicast mode in terms of latency performance. The mean round-trip latency for group mode, in multi-hop and single-hop groups, is 51.62 ms and 38.57 ms, respectively. The unicast mode provides a mean round-trip latency of 152.27 ms and 133.37 ms, respectively. The results indicate that Bluetooth mesh can achieve perfect reliability under unicast mode. The reliability of group mode can be enhanced by configuring more advertising events at the network layer. 

\begin{figure}
\centering
\includegraphics[scale=0.28]{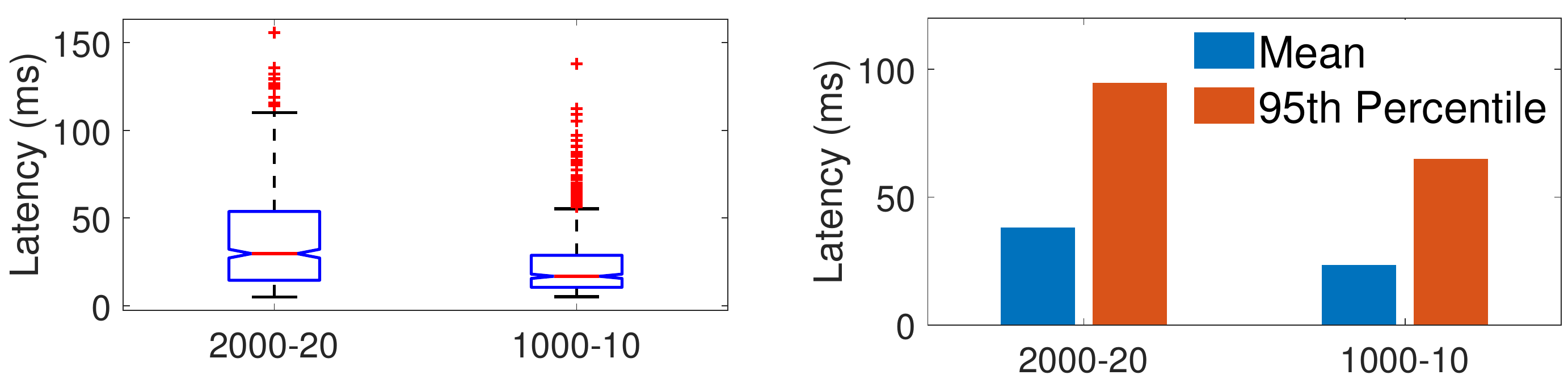}
\caption{Latency performance in many-to-many communication with 7 senders.}
\label{lat_m2m}
\vspace{-1.4em}
\end{figure}
\begin{figure}
\centering
\includegraphics[scale=0.28]{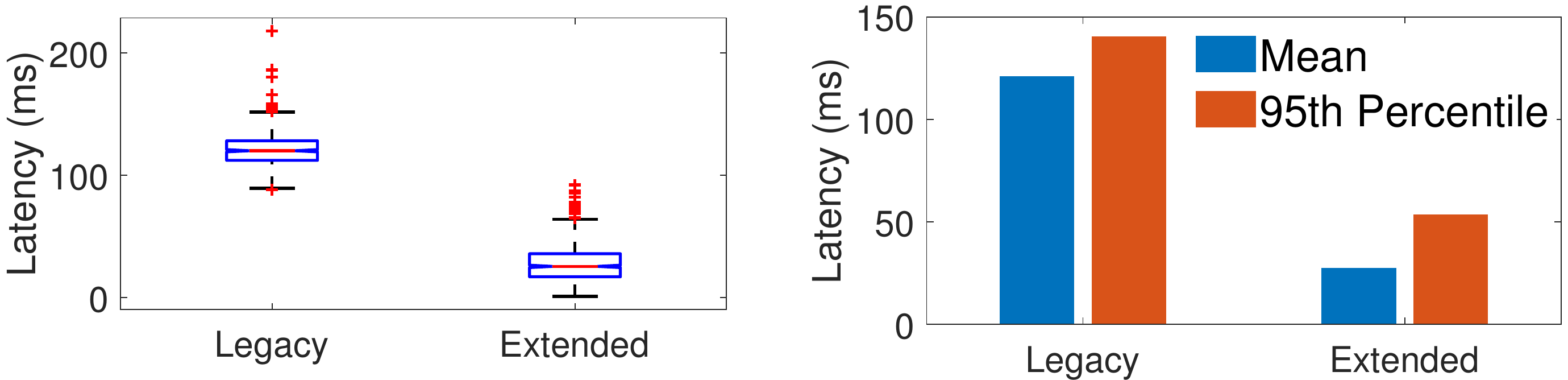}
\caption{Latency performance of legacy and extended advertisements.}
\label{lat_e}
\vspace{-1.4em}
\end{figure}

\subsection{Performance of Many-to-Many Communication Capability}
Many-to-Many communication is crucial for distributed communication and control. We consider a high-traffic many-to-many multi-hop communication scenario wherein 7 concurrent senders transmit to a group of 7 distinct destinations. The results for one-way latency under 100\% reliability are shown in Fig. \ref{lat_m2m}. The box plot captures the statistics of our evaluation. On each box, the central mark is the median, the edges are the 25th/75th percentiles, the whiskers capture the 95\% confidence interval (CI), the outliers are shown by `+', and the notches represent 95\% CI of the median.  We also consider two different configurations of \emph{scanInterval} and \emph{advInterval} parameters, i.e., 2000-20 ms (default) and 1000-10 ms (enhanced). The mean latency under default parameters is 37.89 ms (95th percentile is 94.55 ms). The mean and 95th percentile latency decreases under enhanced configuration to 23.33 ms and 64.77 ms, respectively. Results show the viability of distributed communication/control over Bluetooth mesh.

\subsection{Performance of Bluetooth 5.0 Extended Advertisements}
Extended advertisements  are promising for transmitting bigger payloads in mesh network. We use the Nordic proprietary Instaburst
(https://infocenter.nordicsemi.com/index.jsp) feature for evaluation. It uses a subset of Bluetooth 5.0 extended advertisements with 2 Mbps PHY layer. When enabled, all mesh communication takes place via extended advertisements. The one-way latency results under 100\% reliability for many-to-many communication with 3 concurrent senders and 50 byte messages are shown in Fig. \ref{lat_e}. The mean latency for extended advertisements is 27.28 ms whereas that of legacy advertisements is 120.8 ms. The respective 95th percentiles are 53.39 ms and 140.35 ms. The results show the effectiveness of extended advertisements in handling larger payloads (which can arise in scenarios including exchange of coverage-related and navigation-related information)  with much lower latency.

\subsection{Performance under Mobility}



\begin{figure}
\centering
\begin{subfigure}{0.25\textwidth}
  \centering
  \includegraphics[scale=0.13]{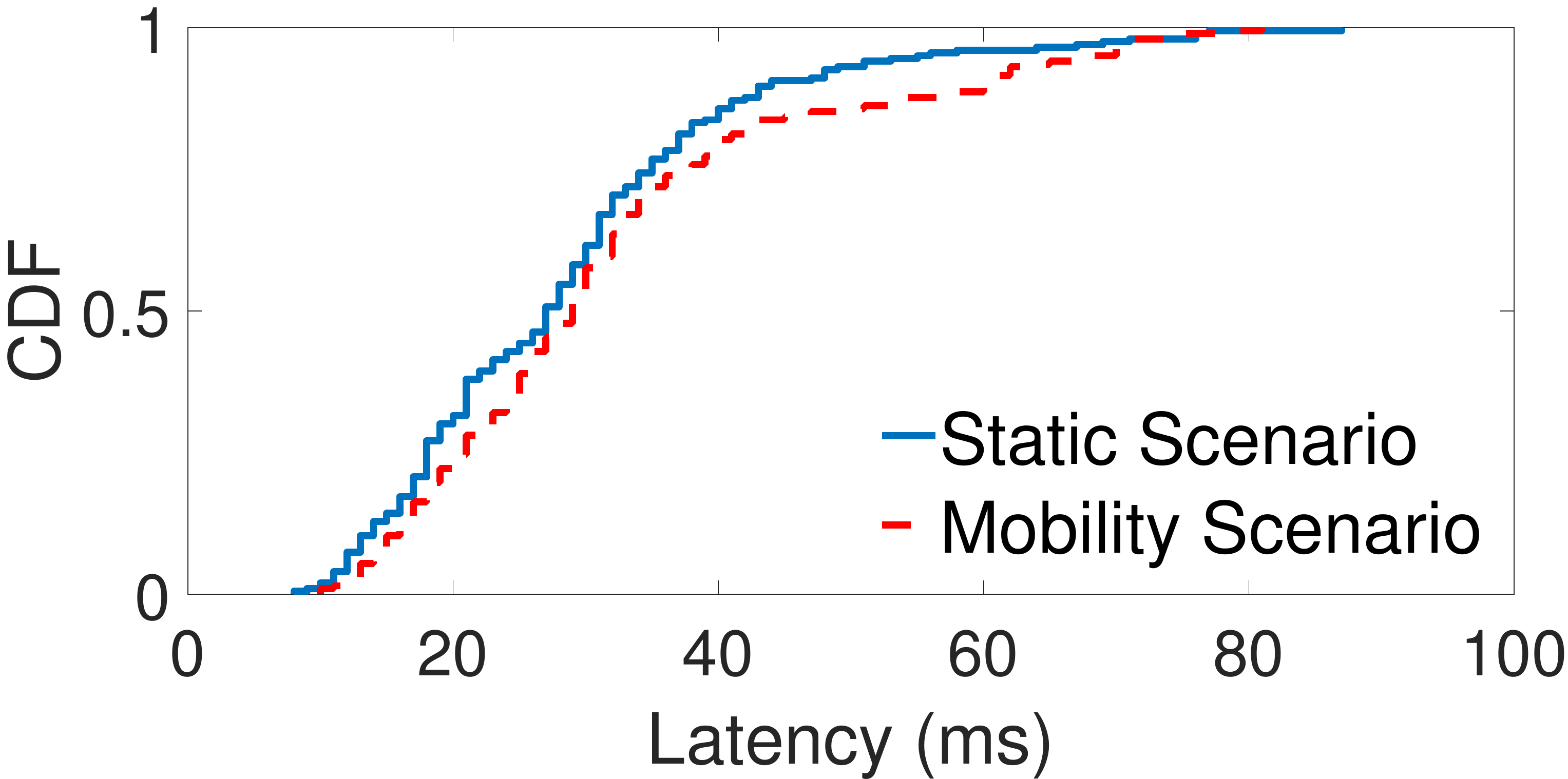}
  \caption{}
  \label{mob_lat}
\end{subfigure}%
\begin{subfigure}{0.25\textwidth}
  \centering
  \includegraphics[scale=0.13]{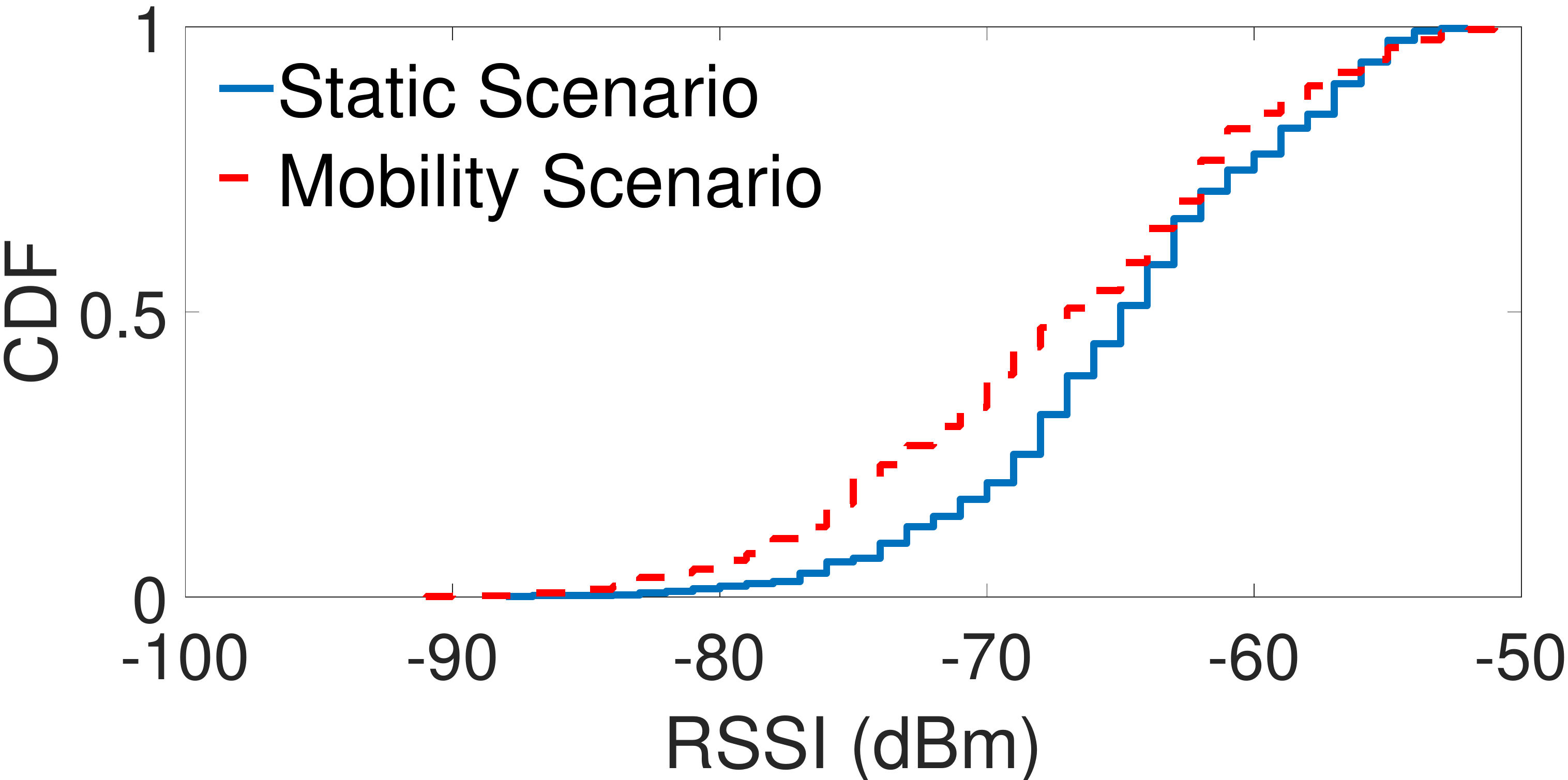}
  \caption{}
  \label{mob_r}
\end{subfigure}
\caption{Latency and RSSI in static and mobility scenarios.}
\label{mob_perf}
\vspace{-1.7em}
\end{figure}

In the second testbed, we introduce mobile nodes. Each mobile node is a GoPiGo3  robot (https://www.dexterindustries.com/gopigo3/) connected to an nRF52840 board. The overall scenario comprises 4 mobile nodes and one controller (nRF52840 board). There are two active mobile nodes: one performs random movement (in the highlighted area) and the other performs controlled (pre-programmed forward/backward) movement on the highlighted path. The remaining mobile nodes are passive. This setup depicts a sparse and dynamic network where the topology changes significantly with  mobility.  We consider a group communication scenario under unicast mode. Fig. \ref{mob_lat} shows the cumulative distribution function (CDF) of round-trip latency for 100\% reliability. We benchmark mobility scenario against a static scenario where no nodes are moving. The results show that latency increases under mobility. The mean latency in mobility scenario is 32.01 ms (95th percentile of 70.5 ms). In static scenario, the mean latency is 28.6 ms (95th percentile of 55.8 ms). The higher latency under mobility is due to degradation in link-level performance, as corroborated by RSSI results in Fig. \ref{mob_r}, which ultimately leads to more retransmissions. However, the results show the viability of successful operation of Bluetooth mesh under mobility.

%
%

\subsection{Comparison with Other Mesh Technologies}
We conduct a qualitative comparison of Bluetooth mesh and other mesh technologies including conventional routing-based\footnote{Techniques which enhance conventional routing-based solutions for mobility-centric operation such as \cite{route_rob2}, \cite{route_rob3}, and \cite{route_rob4}} and recent synchronous flooding\footnote{The term synchronous flooding  broadly refers to a new class of low-power communication protocols that leverage synchronous transmissions which exploit two key physical-layer effects: capture effect and constructive interference. A prominent synchronous flooding protocol is \emph{Glossy} \cite{glossy} which has played a seminal role in development of synchronous flooding protocols. A survey of synchronous flooding protocols is conducted in \cite{Survey_SF}.  Synchronous flooding techniques require network-wide time synchronization and can be realized over different radios including Bluetooth and IEEE 802.15.4. } techniques which are mobility-agnostic. 
The comparison in \tablename~\ref{comp_mesh}  is mainly in terms of  key capabilities for addressing design challenges for ad-hoc logistics as  comparison of latency/reliability metrics does not provide meaningful insights for system-level design. The key points of this comparison are described as follows.

\begin{itemize}
\item Bluetooth mesh and synchronous flooding are transparent to topological changes; hence there is native mobility support. This isn't the case with routing-based techniques which require enhancements for mobility. 

\item Synchronous flooding techniques can support different communication patterns like Bluetooth mesh; however, such techniques require \emph{centralized} scheduling\footnote{In the absence of scheduling, synchronous flooding techniques can only support infrequent traffic which is not suitable for mobile robotic networks.} in order to be scalable. This limits support for heterogeneous traffic, particularly in terms of simultaneously supporting varying  payloads and message frequencies. Routing-based techniques can only support limited number of traffic patterns as dictated by routing trees. 

\item Synchronous flooding and routing solutions need reconfiguration of other nodes in dynamic scenarios. For former, it is schedule reconfiguration (and possible recalibration of network-wide flooding time). For latter, reconfiguration of network state and schedule is necessary. 

\item Synchronous flooding and routing-based techniques are not completely free from centralized management aspects and lack the capability of supporting parallel operation of multiple robotic formations. Hence, these techniques do not offer flexible connectivity like Bluetooth mesh. 

\item Synchronous flooding is prone to a number of security vulnerabilities. This is because of the flooding mechanism which requires all nodes in the network to transmit the exact same message which eliminates the opportunity of using link-level encryption. The entire network can face disruption in case the time synchronizing node is down for any reason. Besides, compromised nodes can be easily added into the network. On the hand, security of routing-based solutions is an implementation-specific issue. 

\item Coverage optimization in routing-based techniques becomes challenging due to network-layer connectivity constraints. For synchronous flooding, coverage optimization becomes challenging as accurate estimate of link-level received power is hard to obtain due to cumulative received signal from multiple nodes. 

\item Synchronous flooding and routing solutions do not offer capabilities like area isolation and seamless addition/removal of nodes from a  formation. Cooperative tasks like formation synthesis and navigation/localization tasks are difficult to handle in conjunction with other tasks. Hence, these technique offer limited support for cooperation-centric and navigation-centric operation.

\end{itemize}

\begin{table}[]
\begin{center}
\caption{Comparison of Capabilities for Mobile Robotic Systems}
\begin{tabular}{lccc}
\toprule
\multicolumn{1}{c}{\textbf{Capability}}                                          & \textbf{\begin{tabular}[c]{@{}c@{}}Bluetooth \\ Mesh\end{tabular}} & \textbf{\begin{tabular}[c]{@{}c@{}}Synchronous\\ Flooding\end{tabular}} & \textbf{\begin{tabular}[c]{@{}c@{}}Routing\\ Solutions\end{tabular}} \\ \hline \midrule
\begin{tabular}[c]{@{}l@{}}Native support \\ for mobility\end{tabular}           & Yes                                                                & Yes                                                                     & No                                                                         \\ \hline
\begin{tabular}[c]{@{}l@{}}Heterogeneous\\ traffic\end{tabular}                  & Yes                                                                & Limited                                                                 & \begin{tabular}[c]{@{}c@{}}Very \\ limited\end{tabular}                    \\ \hline
\begin{tabular}[c]{@{}l@{}}Dynamic \\ environments\end{tabular}                  & Yes                                                                & No                                                                      & No                                                                         \\ \hline
\begin{tabular}[c]{@{}l@{}}Flexible \\ operation\end{tabular}                    & Yes                                                                & No                                                                      & No                                                                         \\ \hline
\begin{tabular}[c]{@{}l@{}}Security \\ vulnerabilities\end{tabular}              & \begin{tabular}[c]{@{}c@{}}Very\\ low\end{tabular}                 & \begin{tabular}[c]{@{}c@{}}Very\\ high\end{tabular}                     & --                                                                         \\ \hline
\begin{tabular}[c]{@{}l@{}}Coverage \\ optimization\end{tabular}                 & Simple                                                             & Hard                                                             & Hard \\ \hline
\begin{tabular}[c]{@{}l@{}}Cooperative\\ operation support\end{tabular}          & Native                                                             & \begin{tabular}[c]{@{}c@{}}Limited \\ support\end{tabular}              & \begin{tabular}[c]{@{}c@{}}Limited\\ support\end{tabular}                  \\ \hline
\begin{tabular}[c]{@{}l@{}}Connect--Navigat. \\co-design\end{tabular} & \begin{tabular}[c]{@{}c@{}}High\\ support\end{tabular}             & \begin{tabular}[c]{@{}c@{}}Limited\\ support\end{tabular}               & \begin{tabular}[c]{@{}c@{}}Limited\\ support\end{tabular}                  \\ \hline
\end{tabular}
\label{comp_mesh}
\end{center}
\vspace{-2.5em}
\end{table}

\section{Concluding Remarks} \label{sect_cr}
The material handling industry is witnessing increased penetration of mobile robotic systems in ad-hoc logistics environments with little/no infrastructure. A team of small and low-cost mobile robots operating in out-of-box and cooperative manner provides a promising solution for these environments. Wireless connectivity plays a crucial role in successful operation of such mobile robotic systems. We have investigated system-level design challenges of mobile robotic systems for ad-hoc logistics and the role of Bluetooth mesh for addressing these challenges. Bluetooth mesh offers unrivalled capabilities for addressing communication, cooperation, coverage, control, security, and navigation/localization challenges. Our real-world evaluation of Bluetooth mesh not only shows its viability for mobile environments but also provides various performance insights. In particular, it demonstrates design flexibility of Bluetooth mesh, its versatility, and near real-time latency performance with perfect reliability. Qualitative comparison also reveals limitations of competing mesh technologies: synchronous flooding and routing-based solutions. 
\vspace{-1.5em}

\section{Acknowledgement}
The author thanks Dominic London, Aleksandar Stanoev, and Victor Marot for contributions to experimental evaluation, and Mahesh Sooriyabandara, Anthony Portelli, Sabine Hauert,  Simon Jones, Tim Farnham, Marius Jurt, Yichao Jin, Usman Raza, and Mahendra Tailor
for partaking in fruitful discussions. 
\vspace{-0.35em}

\bibliographystyle{IEEEtran}
\bibliography{Coop_Rob.bib}

\begin{thebibliography}{10}
\providecommand{\url}[1]{#1}
\csname url@samestyle\endcsname
\providecommand{\newblock}{\relax}
\providecommand{\bibinfo}[2]{#2}
\providecommand{\BIBentrySTDinterwordspacing}{\spaceskip=0pt\relax}
\providecommand{\BIBentryALTinterwordstretchfactor}{4}
\providecommand{\BIBentryALTinterwordspacing}{\spaceskip=\fontdimen2\font plus
\BIBentryALTinterwordstretchfactor\fontdimen3\font minus
  \fontdimen4\font\relax}
\providecommand{\BIBforeignlanguage}[2]{{%
\expandafter\ifx\csname l@#1\endcsname\relax
\typeout{** WARNING: IEEEtran.bst: No hyphenation pattern has been}%
\typeout{** loaded for the language `#1'. Using the pattern for}%
\typeout{** the default language instead.}%
\else
\language=\csname l@#1\endcsname
\fi
#2}}
\providecommand{\BIBdecl}{\relax}
\BIBdecl

\bibitem{report_log_wh}
\BIBentryALTinterwordspacing
IDTechEx, ``{Mobile Robots, Autonomous Vehicles, and Drones in Logistics,
  Warehousing, and Delivery 2020-2040},'' Research Report. [Online]. Available:
  \url{https://tinyurl.com/75ukkyxc}
\BIBentrySTDinterwordspacing

\bibitem{TI_PIEEE}
A.~Aijaz and M.~Sooriyabandara, ``{The Tactile Internet for Industries: A
  Review},'' \emph{Proc. IEEE}, vol. 107, no.~2, pp. 414--435, 2018.

\bibitem{mmh_covid}
\BIBentryALTinterwordspacing
{MMH Staff}, ``{Other Voices: Mobile Robots in the Time of Coronavirus},''
  \emph{Modern Materials Handling}, July 2020. [Online]. Available:
  \url{https://www.mmh.com/article/other_voices_mobile_robots_in_the_time_of_coronavirus}
\BIBentrySTDinterwordspacing

\bibitem{BT_mesh}
{Bluetooth SIG}, ``{Mesh Profile Specification},''
  \url{https://www.bluetooth.com/specifications/mesh-specifications}, Jan.
  2019.

\bibitem{BT_5}
------, ``{Bluetooth Core Specification},''
  \url{https://www.bluetooth.com/specifications/bluetooth-core-specification/},
  Jan. 2019.

\bibitem{MANET_ref}
M.~Conti and S.~Giordano, ``{Mobile Ad hoc Networking: Milestones, Challenges,
  and New Research Directions},'' \emph{IEEE Commun. Mag.}, vol.~52, no.~1, pp.
  85--96, 2014.

\bibitem{mesh_no}
E.~A. Jarchlo, J.~Haxhibeqiri, I.~Moerman, and J.~Hoebeke, ``{To Mesh or not to
  Mesh: Flexible Wireless Indoor Communication Among Mobile Robots in
  Industrial Environments},'' in \emph{Ad-hoc, Mobile, and Wireless
  Networks}.\hskip 1em plus 0.5em minus 0.4em\relax Springer International
  Publishing, 2016, pp. 325--338.

\bibitem{rob_mesh_sensor}
P.~Ghosh, A.~Gasparri, J.~Jin, and B.~Krishnamachari, \emph{{Robotic Wireless
  Sensor Networks}}.\hskip 1em plus 0.5em minus 0.4em\relax Springer
  International Publishing, 2019, pp. 545--595.

\bibitem{route_rob1}
S.~M. Das, Y.~C. Hu, C.~S.~G. Lee, and Y.-H. Lu, ``{Mobility-aware Ad hoc
  Routing Protocols for Networking Mobile Robot Teams},'' \emph{Journal of
  Communications and Networks}, vol.~9, no.~3, pp. 296--311, 2007.

\bibitem{route_rob2}
S.~Maxon and F.~Zhang, ``{Extending a Routing Protocol for Mobile Robot Mesh
  Networking},'' in \emph{IEEE CCTA}, 2017, pp. 37--42.

\bibitem{route_rob3}
N.~Trivedi, B.~Panigrahi, H.~K. Rath, and A.~Pal, ``{Wireless Mesh Routing For
  Indoor Robotic Communications},'' in \emph{ACM IoPARTS}, New York, NY, USA,
  2018, p. 25–30.

\bibitem{wireless_collab}
K.-C. Chen and H.-M. Hung, ``{Wireless Robotic Communication for Collaborative
  Multi-Agent Systems},'' in \emph{IEEE ICC}, 2019, pp. 1--7.

\bibitem{collab_multi}
J.~P. Queralta \emph{et~al.}, ``{Collaborative Multi-Robot Search and Rescue:
  Planning, Coordination, Perception, and Active Vision},'' \emph{IEEE Access},
  vol.~8, pp. 191\,617--191\,643, 2020.

\bibitem{swarm_rob1}
M.~Li \emph{et~al.}, ``{Robot Swarm Communication Networks: Architectures,
  Protocols, and Applications},'' in \emph{IEEE ChinaCom}, 2008, pp. 162--166.

\bibitem{swarm_rob2}
N.~Majcherczyk and C.~Pinciroli, ``{SwarmMesh: A Distributed Data Structure for
  Cooperative Multi-Robot Applications},'' in \emph{IEEE ICRA}, 2020, pp.
  4059--4065.

\bibitem{swarm_rob3}
W.~Li and W.~Shen, ``{Swarm Behavior Control of Mobile Multi-robots with
  Wireless Sensor Networks},'' \emph{Journal of Network and Computer
  Applications}, vol.~34, no.~4, pp. 1398--1407, 2011.

\bibitem{umb_rob}
T.~Farnham \emph{et~al.}, ``{UMBRELLA Collaborative Robotics Testbed and IoT
  Platform},'' in \emph{IEEE CCNC}, 2021, pp. 1--7.

\bibitem{und_perf_BT_mesh}
R.~{Rondón}, A.~{Mahmood}, S.~{Grimaldi}, and M.~{Gidlund}, ``{Understanding
  the Performance of Bluetooth Mesh: Reliability, Delay, and Scalability
  Analysis},'' \emph{IEEE Internet Things J.}, vol.~7, no.~3, pp. 2089--2101,
  2020.

\bibitem{BT_mesh_analysis}
A.~{Hernández-Solana} \emph{et~al.}, ``{Bluetooth Mesh Analysis, Issues, and
  Challenges},'' \emph{IEEE Access}, vol.~8, pp. 53\,784--53\,800, 2020.

\bibitem{exp_BT_mesh_2}
M.~Baert \emph{et~al.}, ``{The Bluetooth Mesh Standard: An Overview and
  Experimental Evaluation},'' \emph{Sensors}, vol.~18, no.~8, 2018.

\bibitem{BT_mesh_overview}
{M. Woolley}, ``{Bluetooth Mesh Networking},''
  \url{https://www.bluetooth.com/wp-content/uploads/2019/03/Mesh-Technology-Overview.pdf/},
  Dec. 2020.

\bibitem{comms_formation}
X.~X. Teh, A.~Aijaz, A.~Portelli, and S.~Jones, ``{Communications-Based
  Formation Control of Mobile Robots: Modeling, Analysis and Performance
  Evaluation},'' in \emph{ACM MSWiM}, 2020, p. 149–153.

\bibitem{coop_nav}
F.~Ducatelle \emph{et~al.}, ``{Cooperative Navigation in Robotic Swarms},''
  \emph{Swarm Intelligence}, vol.~8, pp. 1--33, Mar. 2014.

\bibitem{BT_RDF}
{M. Woolley}, ``{Bluetooth Direction Finding: A Technical Overview},''
  \url{https://www.bluetooth.com/bluetooth-resources/bluetooth-direction-finding/},
  Feb. 2021.

\bibitem{route_rob4}
A.~Elsts \emph{et~al.}, ``{Instant: A TSCH Schedule for Data Collection from
  Mobile Nodes},'' in \emph{ACM EWSN}, 2019, p. 35–46.

\bibitem{glossy}
F.~Ferrari, M.~Zimmerling, L.~Thiele, and O.~Saukh, ``{Efficient Network
  Flooding and Time Synchronization with Glossy},'' in \emph{ACM/IEEE IPSN},
  April 2011, pp. 73--84.

\bibitem{Survey_SF}
M.~Zimmerling, L.~Mottola, and S.~Santini, ``{Synchronous Transmissions in
  Low-Power Wireless: A Survey of Communication Protocols and Network
  Services},'' \emph{ACM Comput. Surv.}, vol.~53, no.~6, Dec. 2020.

\end{thebibliography}
\end{document}